\documentclass[lettersize,journal]{IEEEtran}
\usepackage{amsmath,amsfonts}
\usepackage{algorithmic}

\usepackage{amssymb}
\usepackage{array}
\usepackage[caption=false,font=normalsize,labelfont=sf,textfont=sf]{subfig}
\usepackage{textcomp}
\usepackage{stfloats}
\usepackage{url}
\usepackage{booktabs}
\usepackage{cite}
\usepackage{verbatim}
\usepackage{graphicx}
\usepackage{graphbox}
\usepackage{multirow}
\usepackage{cite,url}
\usepackage{hyperref}
\usepackage{cleveref}
\def\BibTeX{{\rm B\kern-.05em{\sc i\kern-.025em b}\kern-.08em
    T\kern-.1667em\lower.7ex\hbox{E}\kern-.125emX}}
\usepackage{balance}
\begin{document}
\title{NPVForensics: Jointing Non-critical Phonemes and Visemes for Deepfake Detection\thanks{This work was supported in part by the National Key R\&D Program of China(No.2021ZD0112100), National Natural Science Foundation of China (No.U1936212, No.62120106009), and the Beijing Natural Science Foundation (No.4222014). Corresponding author: Rongrong Ni.}
\thanks{Yu Chen, Yang Yu, Rongrong Ni and Yao Zhao are with the Institute of Information Science, Beijing Jiaotong University, Beijing 100044, China, and also with the Beijing Key Laboratory of Advanced Information Science and Network Technology, Beijing 100044, China (e-mail: 21112004@bjtu.edu.cn; 18112012@bjtu.edu.cn; rrni@bjtu.edu.cn; yzhao@bjtu.edu.cn).}
\thanks{Haoliang Li is with the Department of Electrical Engineering, City University of Hong Kong (e-mail: haoliang.li@cityu.edu.hk).}}
 \author{Yu Chen, Yang Yu, Rongrong Ni\*,~\IEEEmembership{Member, IEEE}, Yao Zhao,~\IEEEmembership{Fellow, IEEE}, Haoliang Li}% <-this % stops a space
   \maketitle

\begin{abstract}

Deepfake technologies empowered by deep learning are rapidly evolving, creating new security concerns for society. Existing multimodal detection methods usually capture audio-visual inconsistencies to expose Deepfake videos. 
More seriously, the advanced Deepfake technology realizes the audio-visual calibration of the critical phoneme-viseme regions, achieving a more realistic tampering effect, which brings new challenges.
To address this problem, we propose a novel Deepfake detection method to mine the correlation between Non-critical Phonemes and Visemes, termed NPVForensics. Firstly, we propose the Local Feature Aggregation block with Swin Transformer (LFA-ST) to construct non-critical phoneme-viseme and corresponding facial feature streams effectively. 
Secondly, we design a loss function for the fine-grained motion of the talking face to measure the evolutionary consistency of non-critical phoneme-viseme. 
Next, we design a phoneme-viseme awareness module for cross-modal feature fusion and representation alignment, so that the modality gap can be reduced and the intrinsic complementarity of the two modalities can be better explored.
Finally, a self-supervised pre-training strategy is leveraged to thoroughly learn the audio-visual correspondences in natural videos. In this manner, our model can be easily adapted to the downstream Deepfake datasets with fine-tuning.
Extensive experiments on existing benchmarks demonstrate that the proposed approach outperforms state-of-the-art methods.
\end{abstract}


%%%%%%%%%%%%%
%Introduction
%%%%%%%%%%%%%

\section{Introduction}
\IEEEPARstart{R}{ecently,} face manipulation techniques empowered by deep generative models have made considerable progress \cite{ding2018robust,kwok2021deepfake,abdulreda2022landscape,yamagishi2021asvspoof}, which makes Deepfake media more lifelike. Realistic Deepfake videos are likely to be utilized by attackers for malicious purposes, such as creating and distributing fake news, defaming celebrities, leading to serious security problems. Therefore, there is an urgent demand to develop effective detection methods to reduce the misuse of Deepfake.

\begin{figure}[]
  \centering
  \includegraphics[width=\linewidth]{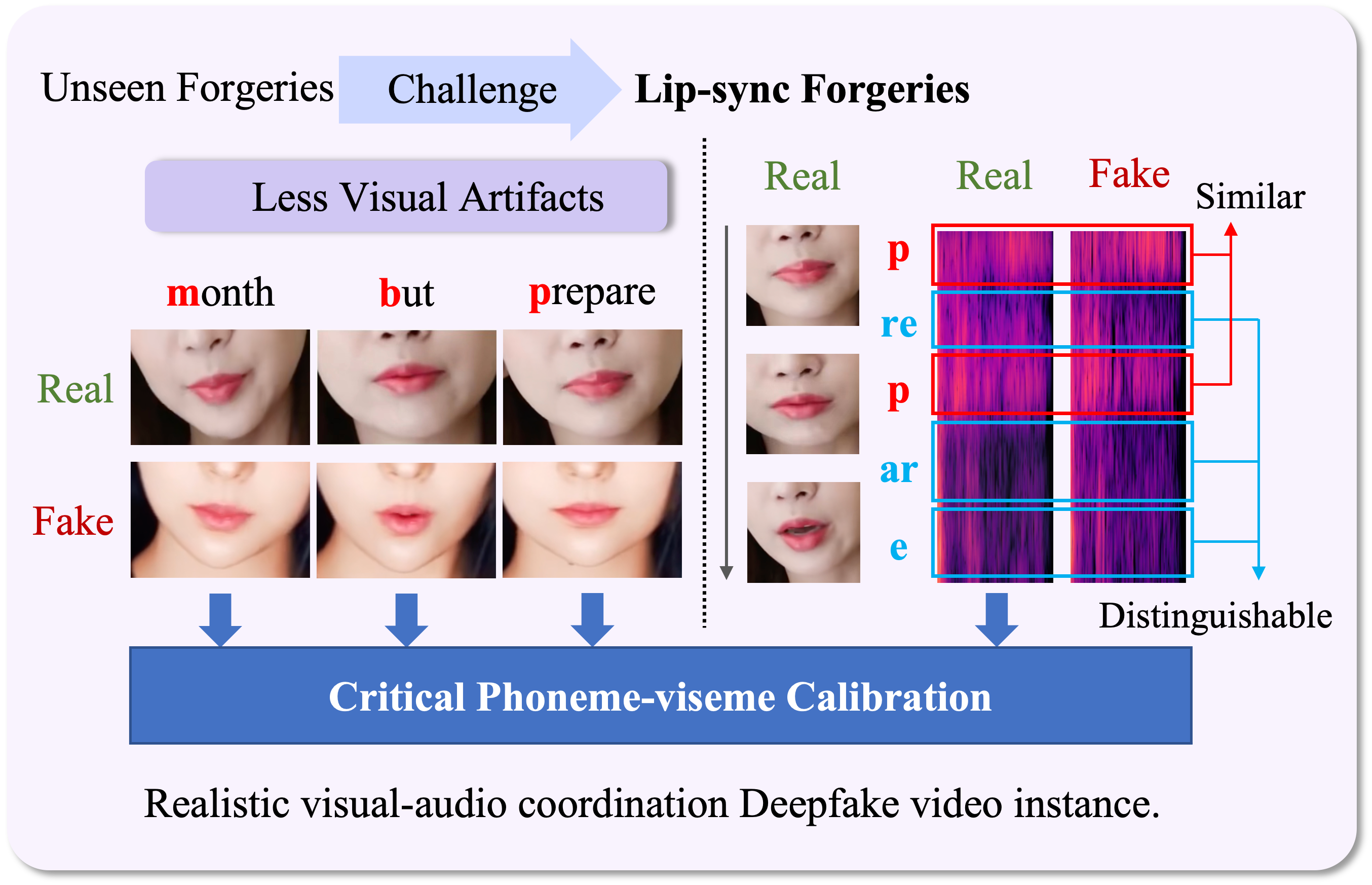}
  \caption{This figure displays a challenging Lip-Sync forgery video example. On the left, the first row of video frames is real, while the second row is Deepfake. We observe that the forger has calibrated the phoneme-viseme pairs in critical phoneme regions like `m', `b' and `p'. Therefore, the difference between the real and fake visemes is small in the critical phoneme regions. The right column of video frames is real,  uttering the word `prepare', corresponding to the first column of Mel-spectrograms. The second column of Mel-spectrograms is generated by $text\ to\ speech$ (TTS) algorithm\cite{jia2018transfer}. We can observe that the two columns of Mel-spectrograms are similar at the critical phoneme `p' (marked in red), while they are easily distinguishable at the non-critical phonemes `re', `ar' and `e' (marked in blue). The phoneme-viseme correlations in non-critical regions are what we aim to capture in this work.}
  \label{fig:challenges}
  \vspace{-0.4cm}
\end{figure}

So far, some Deepfake detection methods have achieved good performance on specific datasets \cite{chen2015automatic,feng2016motion,d2018patchmatch,kong2022detect,masi2020two,xie2020facial,yu2020mining,hu2021detecting,aloraini2020sequential}. These methods are dedicated to mining visual cues, such as local textures \cite{chen2015automatic,d2018patchmatch}, noise characteristics \cite{masi2020two,kong2022detect}, frequency information \cite{yu2020mining} etc. LipForensics \cite{haliassos2021lips} utilizes pretrained lip-reading networks to refine the model and learn embeddings more sensitive to lip motions. 
In our previous work\cite{li2022artifacts}, we proposed an artifacts-disentangled adversarial learning framework to achieve accurate Deepfake detection by disentangling the artifacts from irrelevant information. However, the grim reality is that some forgery methods have started to tamper with audio and visual information\cite{khalid2021evaluation}, which has more harmful effects on information dissemination. Therefore, the above approaches focusing only on visual forgery cue mining have significant limitations.

To address this problem, more and more researchers have started focusing on the role of audio in Deepfake detection. The papers \cite{chugh2020not,agarwal2020detecting} perform Deepfake detection based on measuring the consistency between two modalities. However, these two approaches do not take into account the complementary information of audio-visual modalities. 
The human brain integrates the received audio-visual information to collaborate and complement each other\cite{bedi2021multi}. Inspired by human cognition, integrating audio-visual data, i.e., exploring audio-visual complementarity, would benefit the information processing capability and model robustness of the Deepfake detection. The consistency of audio-visual modalities provides the basis for multimodal Deepfake detection, and the intrinsic complementarity between the two makes it possible to enhance the signal of another modality with the help of one of the modalities. VFD \cite{cheng2022voice} considers homogeneity between faces and voices in natural videos. However, the model fails to detect tampered data without identity change, such as removing glasses. %RealForensics\cite{haliassos2022leveraging} exploits the natural correspondence between the visual and audio modalities in real videos to learn temporally dense video representations in a self-supervised cross-modal manner.
AVoiD-DF\cite{yang2023avoid} designs a dual-stream temporal-spatial encoder and a multimodal joint decoder for joint learning of intrinsic relations. The paper\cite{zhou2021joint} proposes a joint audio-visual detection framework, which utilizes independent audio stream, visual stream and synchronized stream for prediction. Neither of the above two works considers the modalities gap of audio-visual features, so even real videos may be misclassified due to the semantic gap. 
More importantly, audio-visual forgery techniques are evolving rapidly. The new forgery methods \cite{hegde2022lip} can achieve more realistic visual-audio coordination based on the correspondence between critical phonemes and visemes in words, as shown in Fig.~\ref{fig:challenges}. Specifically, this new type of Deepfake technology can calibrate the visual state of the lips when the phonemes `m', `b' and `p' are pronounced. Existing approaches could not address the problem of Deepfake video detection with fine-grained manipulation at the word and phoneme level. To this end, it is a new challenge and an issue that requires urgent resolution.

 Motivated by the above analysis, we discovered that these manipulated methods are based on critical phoneme-viseme to perform audio-visual correspondence. The difference between the real video and the manipulated video is substantially reduced in the critical phoneme-viseme region calibrated by the forger. Thus, this paper focuses on the non-critical phoneme-viseme regions that forgers cannot perfectly reshape. Therefore, we propose a novel Deepfake detection method to mine the correlation of Non-critical Phonemes and Visemes, termed NPVForensics. 
 Firstly, considering that the features in the non-critical phoneme-viseme region are relatively subtle and weakly correlated (compared to the critical phoneme-viseme), we design a Local Feature Aggregation block with Swin Transformer (LFA-ST) to extract the filtered non-critical phonemes and its corresponding visemes and facial stream features. Particularly, LFA-ST not only integrates the local neighboring token signals, but also captures long-range dependencies to obtain global information and improve performance. It facilitates the analysis of the evolutionary consistency of non-critical phoneme-viseme.
Secondly, we propose the non-critical Phoneme-Viseme Awareness Module (PVAM) for joint audio-visual detection. PVAM contains two sub-modules: Cross Attentional Fusion Module (CAFM) and Co-correlation Guided Representation Alignment (CGRA). It can reduce the modality gap and better explore the intrinsic complementarity of non-critical phoneme and viseme, contributing to Deepfake detection. 
Finally, inspired by SST\cite{zhao2022self}, we perform a self-supervised pre-training strategy to obtain a comprehensive representation of non-critical phonemes and visemes. The reason is that this work demonstrates that self-supervised pre-training on relatively large number of real videos can be adequate for learning audio-visual correspondences. In this way, our model can be easily adapted to downstream Deepfake datasets with fine-tuning.

Our contributions are as follows:

1) We propose a novel detection method for joint non-critical phonemes and visemes for Deepfake video with realistic visual-audio forgery, which exploits the correlation between phonemes at non-critical regions in audio utterances and lip motions (visemes).

2) To fully explore the evolutionary inconsistency of non-critical phoneme-viseme, we propose the Local Feature Aggregation Swin Transformer (LFA-ST) to mine subtle features. The integration of local aggregation modules in Swin Transformer ensures maximum retention of global and local information of non-critical phoneme-viseme.

3) To simulate the intrinsic complementary audio-visual relationship, we designed a Phoneme-viseme Awareness module (PVAM) to perform cross-modal fusion of phoneme and viseme features. In particular, the Co-correlation Guided Representation Alignment (CGRA) part can align heterogeneous audio-visual data and reduce the modality gap.

%%%%%%%%%%%%%%
%Related work
%%%%%%%%%%%%%%
\begin{table}[t]
	\centering
	\caption{Description of the 15 critical phonemes targeted by forgers with six classes of corresponding visemes.}
	\label{tablepv}
	\small
	\renewcommand{\arraystretch}{1.2} % default is 1.0
	\begin{tabular}{llll}
		\toprule
		Classes & Viseme & Phoneme & Examples of Words \\
		\midrule
		1  & \includegraphics[align=c]{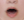} & ay, ah         & but            \\[5pt]
		2  & \includegraphics[align=c]{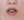} & ey, eh         & bet            \\[5pt]
		3  & \includegraphics[align=c]{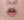} & er             & bird           \\[5pt]
		4  & \includegraphics[align=c]{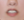} & ch, sh, jh, zh & chase, joke    \\[5pt]
		5  & \includegraphics[align=c]{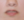} & f, v           & finally        \\[5pt]
		6  & \includegraphics[align=c]{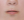} & m, b, p, em    & moment, parent \\
		\bottomrule
	\end{tabular}

\end{table}

\section{Related Work}

In this section, we provide a brief review of Deepfakes with Phoneme-Viseme Coordination, and then we introduce the development of Deepfake detection.

\subsection{Deepfake with Phoneme-Viseme Coordination}

Phonemes are perceptually distinct sound units in speech. A viseme is the visual counterpart of that phoneme and corresponds to the lips shape required to pronounce it. The paper \cite{chen2022transformer} compared phonemes and visemes, listing critical phonemes and corresponding visemes. Table~\ref{tablepv} shows a subset of six visemes for phonemes (a viseme may correspond to more than one phoneme). For example, in order to pronounce the phonemes `m'(mother), `b'(brother) and `p'(parent), a person needs to pronounce the phonemes with the lips completely closed. While the specific shapes of various phonemes may depend on other phonological features such as emphasis or volume, critical phoneme groups such as `m', `b' and `p' always correspond to specific lip shapes. With that in mind, the paper \cite{agarwal2020detecting} performs Deepfake detection based on whether the lips are completely closed at the phoneme `m', `b' and `p'. In contrast, the forger will reshape those critical phonemes with high certainty \cite{vajpayee2023simple}, like the Text-to-Video (T2V) dataset\cite{fried2019text}, containing respective synthesized lip-synced fake audios, where the critical phoneme positions correspond well to the viseme.

Therefore, forgery datasets with challenging critical phoneme-viseme matching cannot be solved by these critical phonemes alone. For those neglected non-critical phonemes, there is a significant `forgery loophole'. Based on the above observations, we propose a novel joint non-critical phoneme-viseme detection method to perform Deepfake detection in  non-critical phoneme-viseme regions that forgers cannot perfectly reshape.

\subsection{Deepfake Detection}

Currently, some researchers focus on mining visual cues that can identify Deepfakes, such as frequency-aware cues \cite{yu2020mining}, anomalous artifacts in the color domain \cite{xia2022towards}, and anomalous parsing features from facial recognition networks. Cao et al. \cite{cao2022end} used pixel-level segmentation mapping methods for Deepfake detection. However, these methods resort to specific forgery patterns of manipulation techniques in the dataset. Recent work can be summarized in two solutions: data enhancement \cite{chen2022self} or capturing general clues\cite{li2020face} to boost the generalization ability. Method \cite{chen2022self} can uncover significant differences between real and fake media. Nevertheless, building such datasets is costly. Paper \cite{shiohara2022detecting} detects blending boundaries between inner faces and outer faces. However, these detection methods are sensitive to post-processing operations, such as compression and blurring. Thus, some studies mine robust high-level semantic features. For example, LipForensics \cite{haliassos2021lips} uses pretrained lipreading networks to refine the model and learn embeddings more sensitive to lip motions. However, unlike our proposed method, none of the above detection methods utilizes audio information. The lip feature extractor is pre-trained and supervised on Lipreading in the Wild (LRW) dataset \cite{2016Lip}, which requires a high annotation cost. 

It is noteworthy that additional audio-visual information can improve perceptual decision-making compared to single visual information. During the audio-visual process, information from different modalities contributes to the rapid accumulation of sensory evidence, accelerating and correcting the final decision. Paper \cite{zhao2022self} performed the collaborative learning of audio-visual models under a self-supervised strategy. Visual and audio can provide semantic information from different perspectives and thus complement each other. To this end, paper\cite{zhou2021joint,haliassos2021lips,zhao2022self} trained their models with these two patterns to learn rich representations. The paper\cite{agarwal2020detecting} performs Deepfake detection based on whether the lips are completely closed at the critical phonemes. However, with the continuous development of lip-sync technology, forgers reshape critical phonemes and visemes in words to achieve a visually consistent effect. Therefore, fine-grained analysis of audio-visual features contributes to Deepfake detection. In this paper, we observe that due to the complexity of lip motions, forgers are unable to reshape all phonemes, and there is still audio-visual dissonance at other locations excluding critical phonemes. That motivates our research to perform Deepfake detection from the perspective of jointing audio-visual, to learn the correspondence between non-critical phonemes and visemes in real videos, and to explore the evolutionary consistency and intrinsic complementary between non-critical phonemes and visemes temporally. Moreover, our method improves generalization and robustness and is more cost-efficient in data annotation.

%%%%%%%%%%
%Method
%%%%%%%%%%

 \begin{figure*}[t]
  \centering
  \includegraphics[width=0.8\linewidth]{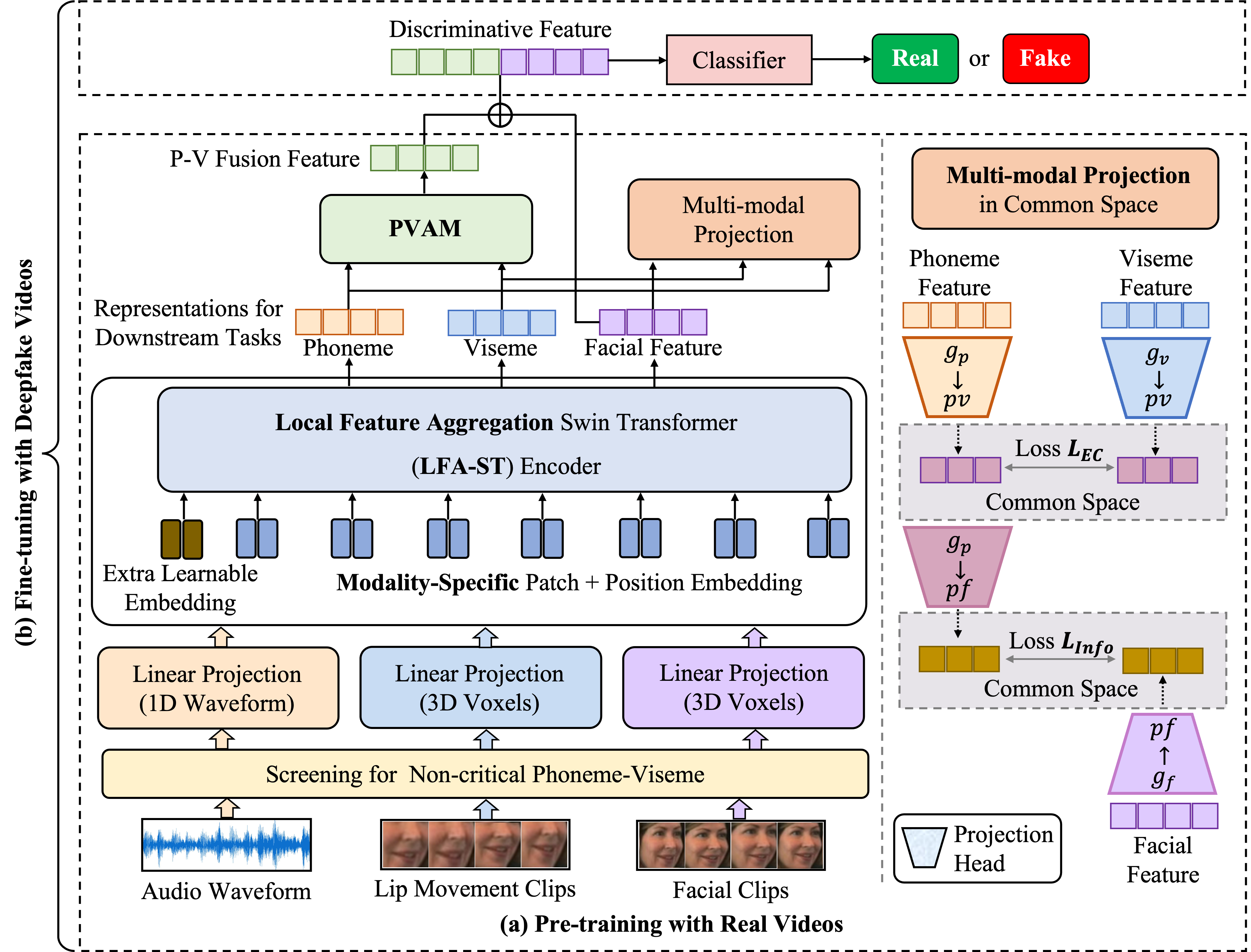}
  \caption{NPVForensics Pipeline. NPVForensics is organized into two stages. The first stage is the pre-training stage. We screen the audio waveform, lip motion stream, and face stream of the real video for non-critical phoneme-viseme, then input them into Swin Transformer Encoder with Local Feature Aggregation module to extract sufficient representations. Then the three streams are fed into Multi-modal Projection Head to calculate the Evolutionary Consistency Loss (EC Loss). The Multi-modal Common Space Projection is shown in the right column. In the meantime, the phoneme and viseme features are fed into the Phoneme-viseme Awareness Module (PVAM) to perform cross-modal fusion and representation alignment. The second stage is the fine-tuning stage. We utilize the Deepfake datasets for fine-tuning. Then we concatenate the fusion features with the facial features as discriminative features for classification.}
   \label{fig:method}
\end{figure*}

\section{Method}

 Most Deepfake generation methods are unable to perform phoneme-viseme fitting on all regions but only on critical phonemes. Thus, non-critical phoneme-viseme inconsistency may be a common feature of Deepfake videos, and such high-level semantic features are robust to various post-processing of videos. This section illustrates the proposed Deepfake detection framework based on self-supervised learning with jointly non-critical phonemes and viesmes, termed NPVForensics. The pipeline of the framework is shown in Fig.~\ref{fig:method}. Our research focuses on mining the correlation between non-critical phonemes and visemes representations. Firstly, we input the non-critical phonemes, visemes, and face streams of the real videos into separate backbone Local Feature Aggregation Swin Transformer (LFA-ST) with modality-specific weights to extract relatively subtle features between non-critical phoneme-viseme (see Section~\ref{feature extraction}). Secondly, we designed EC loss to measure the evolutionary consistency of non-critical phonemes and visemes (see Section~\ref{Evolutionary Consistency Mining}). Then, in order to reduce the modality gap and better investigate the intrinsic complementarity of two modalities, we introduce the non-critical phoneme-viseme awareness module (PVAM), which includes two steps of cross-modal feature fusion and feature alignment (see Section~\ref{NPVAM}). Finally, the self-supervised pre-training strategy and fine-tuning protocol are presented (see Section~\ref{Pre-train}).

\subsection{Audio-visual Feature Extraction}
\label{feature extraction}

We first extract the positions of all phonemes. Google's Speech-to-Text API automatically transcribes the audio tracks associated with the video and then aligns them to the audio using P2FA\cite{rubin2013content}. The above steps generate a series of phonemes as well as its start and end times in the input audio/video. We screen out 15 critical phonemes (including phoneme combinations) shown in Table~\ref{tablepv} to filter non-critical phonemes and its start and end times.

\begin{figure*}[h]
  \centering
  \includegraphics[width=1.0\linewidth]{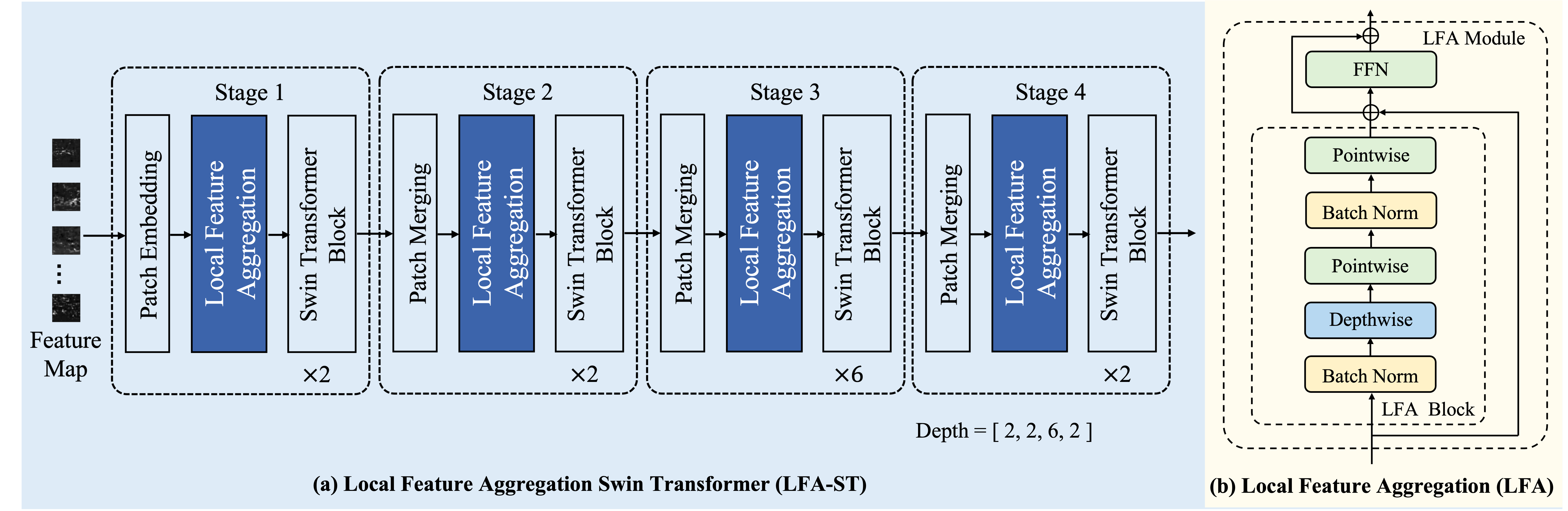}
  \caption{The overall architecture of Local Feature Aggregation Swin Transformer (LFA-ST). (a) shows the embedding position of the LFA module in the Swin Transformer. (b) shows the details of the LFA module.}
  \label{fig:LFA-ST}
  \vspace{-0.3cm}
\end{figure*}

In order to obtain comprehensive and sufficient features of non-critical phonemes, visemes and facial streams, we introduced the Swin Transformer \cite{liu2022video} as a feature encoder in Deepfake detection. %Swin Transformer shifted window mechanism relocates edge windows to the position related to its other features so as to address the issue of losing local information between neighboring windows due to window cutting, and to ensure that global information of each image is acquired and information interaction between different windows is achieved. 
That is advantageous for preserving the global features of the non-critical phoneme, viseme, and facial streams. To maximize the retention of global and local information, we designed a Local Feature Aggregation (LFA) module and integrated it into the structure of the Swin Transformer, as shown in Fig.~\ref{fig:LFA-ST}(a). The LFA block consists of lightweight Depthwise and Pointwise Convolutions that can aggregate local features from feature maps and also can integrate local aggregation of local approximate token signals in each stage of the Transformer, as shown in Fig.~\ref{fig:LFA-ST}(b). 

The LFA block consists of two BatchNorms, two $1\times1$ Pointwise Convolutions and one $3\times3$ Depthwise Convolution. Using effective Depthwise Convolution, local information is collected from neighboring tokens (each token corresponds to a specific patch) to maximize the extraction of detailed cues for non-critical phoneme and viseme features. %Depthwise Convolution does not degrade the performance of the model while reducing the parameters. 
Pointwise Convolution aggregates the information between each channel. %In the LFA module, the BatchNorm layer not only boosts the training speed and convergence process, but also slightly improves the classification results. 
In addition, the method is easy to adjust the reference process, requires less initialization, and can use a larger learning rate. %(We also conducted experiments on Pointwise Convolution followed by Depthwise Convolution, and the difference between the two is not significant, depending on the stacking order of the experimental setup)

For each patch, the information is aggregated in a local window of size $k \times k$ using Pointwise ($1 \times 1$) and Depthwise ($3 \times 3$) convolution. The input signal is sequentially subjected to BatchNorm, Depthwise Convolution and Pointwise convolution operations. It is shown in Equation \eqref{eq:1}.
\begin{equation}\label{eq:1}
  \operatorname{LFA}(\boldsymbol{X}_{in}) = \operatorname{PW}(\operatorname{DW}(\operatorname{BN}(\boldsymbol{X}_{in})).
\end{equation}

Since the Depthwise Convolution operation independently convolves each channel of the input layer, it does not effectively use the feature information of different channels at the same spatial location. Therefore, these feature maps need to be combined by Pointwise Convolution to generate a new feature map. It is shown in Equation \eqref{eq:2}.
\begin{equation}\label{eq:2}
  \operatorname{LFA}(\boldsymbol {X}_{mid}) = \operatorname{PW}(\operatorname{BN}(\operatorname{LFA}(\boldsymbol{X}_{in}))).
\end{equation}

Then it enters the FFN module, and the process is shown in Equation \eqref{eq:3}.
\begin{equation}\label{eq:3}
  \boldsymbol {X}_{out} = \operatorname{FFN}(\operatorname{LFA}(\boldsymbol{X}_{mid})).
\end{equation}

\begin{figure}[t]
  \centering
  \includegraphics[width=0.3\linewidth]{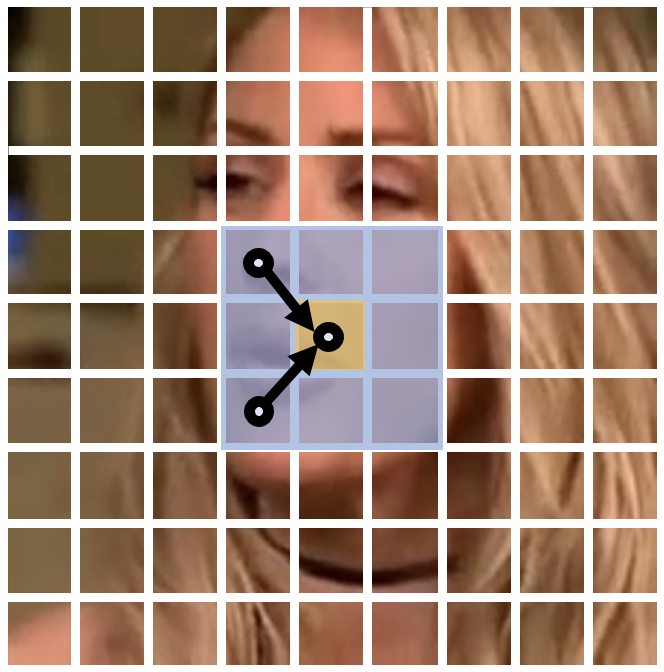}
  \caption{The local information from neighbor tokens within the blue area is first aggregated to the target token.}
  \label{fig:LFA}
  \vspace{-0.3cm}
\end{figure}

%The 3D shifted windows of Swin Transformer are shown in Fig.~\ref{fig:ST shift}. Fig.~\ref{fig:ST shift}(b) demonstrates the role of LFA aggregated neighbor tokens. It is worth noting that LFA module only integrates neighboring Token signals from local sources, whereas with Swin Transformer remote dependencies can be captured to obtain globally information of the input signal and improve performance. 
The schematic diagram of the LFA module aggregating local features is shown in Fig.~\ref{fig:LFA}.

\subsection{Phoneme-viseme Evolutionary Consistency Mining}
\label{Evolutionary Consistency Mining}

LFA-ST operates on each of the three input signals. The audio input is in the form of a waveform, and the visual modal input consists of 3 channels of RGB pixels from a video frame. We first define a modality-specific Tokenization layer, which takes the original signal as input and returns a series of vectors to be fed into LFA-ST. %In addition, each modality has its own position embedding, which injects the order of the tokens into the Swin Transformer. 
The original audio waveform is a one-dimensional input of length $T^{'}$, after phoneme sieving at critical positions. We divide it into $\lceil T^{'}/t^{'}\rceil$ segments, each of which contains $t^{'}$ waveform amplitudes. We apply linear projections with learnable weights  
$\boldsymbol{W}_{p}\in\mathbb{R}^{t^{'}\times d}$ to all elements in the audio patch to derive a d-dimensional vector representation. We employ $\lceil T^{'}/t^{'}\rceil$ learnable embedding to encode the position of each waveform segment. Similarly, video clips of size $T \times H \times W$ corresponding to non-critical phonemes are divided into a series of $\lceil T/t\rceil\cdot\lceil H/h\rceil\cdot\lceil W/w\rceil$ patches where each patch contains $T \times H \times W \times 3$ voxels. We apply linear projections to all voxels in each patch to obtain a d-dimensional vector representation. The projections of visemes and facial streams are performed by the learnable weights $\boldsymbol{W}_{v}\in\mathbb{R}^{t\cdot h\cdot w\cdot \times 3}$ and $\boldsymbol{W}_{f}\in\mathbb{R}^{t\cdot h\cdot w\cdot \times 3}$, respectively.

We utilize common space projection and contrast learning in common space to train our network. More specifically, given a $\langle phonemes, visemes, facial\ stream\rangle$ triplet, we define a semantically hierarchical common space mapping to compare $\langle phonemes, visemes\rangle$ pairs directly as well as $\langle phonemes, facial\ stream\rangle$ pairs by cosine similarity. To achieve this, we define multi-level projections. As shown in the multimodal common space mapping in Fig.~\ref{fig:method}, where $g_{p}\to pv$ and $g_{v}\to pv$ are projection heads that map the outputs of the Phoneme Transformer and the Viseme Transformer, to the phoneme-viseme common space $S_{pv}$, respectively. Similarly, $g_{f}\to pf$ and $g_{p}\to pf$ project the output of the Facial Stream Transformer and the video embedding in the $S_{pv}$ space to the phoneme-facial stream common space $S_{pf}$. The main intuition behind that hierarchy is that different modalities have different levels of semantic granularity. Thus we should treat them as inductive biases in the projection of the common space. We use a linear projection for $g_{p}\to pv(\cdot)$ and $g_{p}\to pf(\cdot)$, and a two-layer projection with ReLU for $g_{v}\to pv(\cdot)$ and $g_{f}\to pf(\cdot)$. Batch normalization is used after each linear layer to simplify the training.  

%Considering the semantic gap of the audio-visual features, the contrast loss still exists even for real videos. 
For the purpose of reducing the false positive rate, we perform a segment-wise `fine-grained' analysis of the phoneme features and the viseme features. It is observed that in Deepfake videos, there are inconsistencies in each segment. Calculating the total sum of inconsistencies during evolution, the faked video is much larger than the real one. Therefore, we evaluate the cumulative loss of phoneme-viseme over time to obtain the total phoneme-viseme evolutionary consistency loss (${EC}$ Loss). Let $h_{\theta}$ be the distance between two features, which is obtained by using the feature representation of non-parametric softmax.
\begin{equation}\label{eq:4}
  h_{\theta}(\boldsymbol{X}_{pi} \cdot \boldsymbol{X}_{vi}) = \exp\left( \frac{\boldsymbol{X}_{pi} \cdot \boldsymbol{X}_{vi}}{\|\boldsymbol{X}_{pi}\| \cdot \|\boldsymbol{X}_{vi}\|} \cdot \frac{1}{\tau} \right).
\end{equation}

\noindent
where $\tau \in \mathbb{R}^{+}$ is a scalar temperature parameter.

The evolutionary consistency contrast loss for each segment is defined as shown in Equation \eqref{eq:5}.
\begin{equation}\label{eq:5}
  L_{{EC}_{i}} = L_{{EC}_{i}}^{p} + L_{{EC}_{i}}^{v}.
\end{equation}

The contrast loss of audio modality is:

\begin{equation}\label{eq:6}
  L_{{EC}_{i}}^{a} = -\log
  \resizebox{0.8\width}{!}{$\displaystyle\frac{h_{\theta}(\boldsymbol{X}_{pi } \cdot \boldsymbol{X}_{vi})}{\sum_{j \in I} h_{\theta}(\boldsymbol{X}_{pi} \cdot \boldsymbol{X}_{vj}) + \sum_{j \in \boldsymbol{A}{(i)}} h_{\theta}(\boldsymbol{X}_{pi} \cdot \boldsymbol{X}_{vj})}$}.
\end{equation}

\noindent
where, $i\epsilon I \equiv \{1,\cdots ,N\}$ and $\boldsymbol{A}(i) \equiv I \setminus \{i\}$. Similarly, the contrast loss of the viseme stream is given by Equation \eqref{eq:7}.

\begin{equation}\label{eq:7}
  L_{{EC}_{i}}^{l} = -\log
  \resizebox{0.8\width}{!}{$\displaystyle\frac{h_{\theta}(\boldsymbol{X}_{vi} \cdot \boldsymbol{X}_{pi})}{\sum_{j \in I} h_{\theta}(\boldsymbol{X}_{vi} \cdot \boldsymbol{X}_{pj}) + \sum_{j \in \boldsymbol{A}{(i)}} h_{\theta}(\boldsymbol{X}_{vi} \cdot \boldsymbol{X}_{pj})}$}.
\end{equation}

Thus, the total EC Loss is calculated as:
\begin{equation}\label{eq:8}
  L_{EC} = \sum_{i\epsilon I} L_{{EC}_{i}}.
\end{equation}

\subsection{Non-critical Phoneme-Viseme Awareness Module}
\label{NPVAM}

\begin{figure*}
  \centering
  \includegraphics[width=0.8\linewidth]{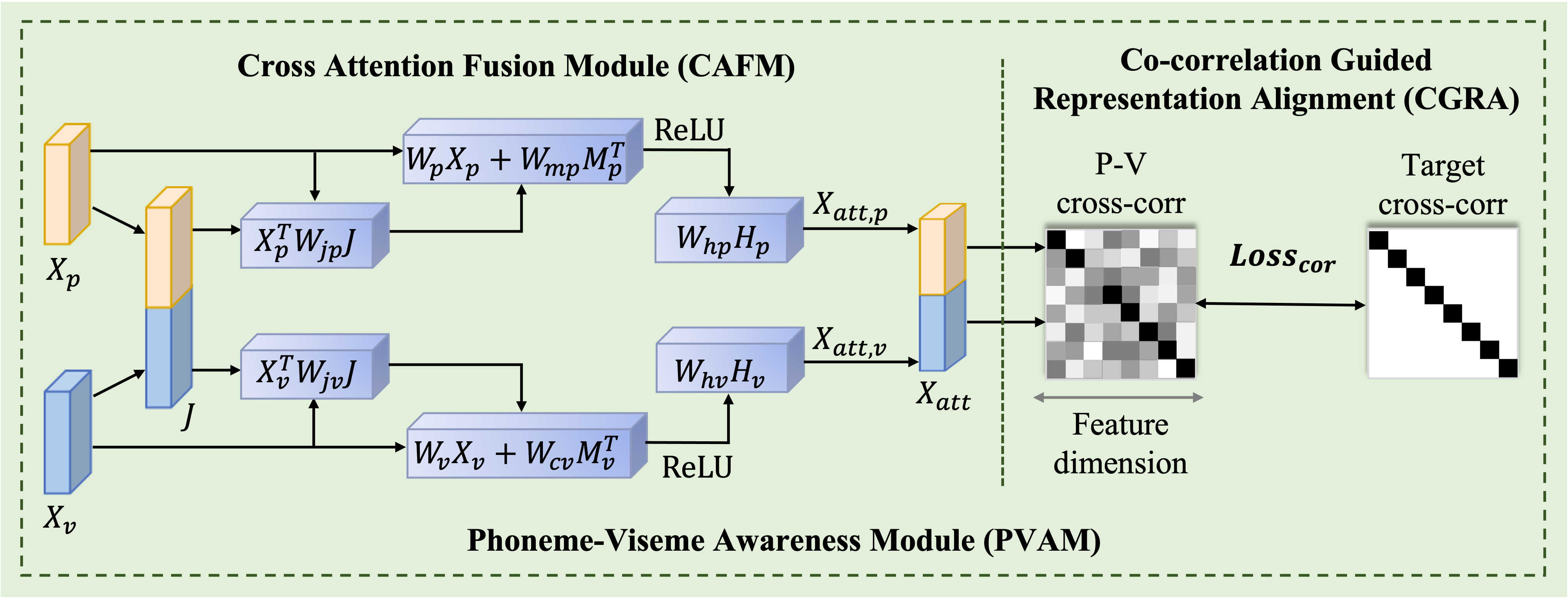}
  \caption{The framework of the Non-critical Phoneme-Viseme Awareness Module (PVAM). The left part shows the Cross Attention Fusion Module (CAFM). Moreover, the right part shows the procedure of Co-correlation Guided Representation Alignment (CGRA).}
  \label{fig:NPVM}
  \vspace{-0.2cm}
\end{figure*}

Generally, the visual modality carries more information relevant to the Deepfake detection task for a given video sequence. In contrast, the phoneme modality is more suitable as a piece of complementary information to the viseme modality. %We need to effectively capture the intrinsic complementary relationship between them. 
To reliably combine these modalities, we resort to a fusion mechanism based on cross attention, where the features of each modality focus on itself and the other modalities, contributing to capturing the semantic inter-modal relations between P (Phoneme) and V (Viseme). We drew the joint modeling of intra and inter-modal relationships into the Non-critical Phoneme-Viseme Awareness Module (PVAM). The phoneme features and the corresponding viseme features are fed into the Cross Attention Fusion Module (CAFM) to explore the intrinsic complementarity between them. To align the heterogeneous audio-visual patterns, we add a Co-correlation Guided Representation Alignment (CGRA) step after CAFM. The proposed model is shown in Fig.~\ref{fig:NPVM}.

\noindent
\textbf{Cross Attention Fusion Module.} Let $\boldsymbol{X}_{p}$ and $\boldsymbol{X}_{v}$ denote the phoneme features and viseme features of the input video respectively, where $\boldsymbol{X}_{p} = \{\boldsymbol{X}_{p1},\boldsymbol{X}_{p2}, \cdots, \boldsymbol{X}_{pn}\} \in \mathbb{R}^{d_{p} \times S}$ and $\boldsymbol{X}_{v} = \{\boldsymbol{X}_{v1},\boldsymbol{X}_{v2}, \cdots, \boldsymbol{X}_{vn}\} \in \mathbb{R}^{d_{v} \times S}$. $S$ denotes the sequence length, $d_{p}$ and $d_{v}$ denote the dimensions of the phoneme features and the viseme features.

As shown in Fig.~\ref{fig:NPVM}, the joint representation of the P-V features is $\boldsymbol{J}$, which is obtained by concatenating the feature vectors of $\boldsymbol{X}_{p}$ and $\boldsymbol{X}_{v}$. The joint correlation matrix $\boldsymbol{M}_{p}$ across the features $\boldsymbol{X}_{p}$ and $\boldsymbol{J}$ are given by Equation \eqref{eq:9}.
\begin{equation}\label{eq:9}
 \boldsymbol{M}_{p}= \tanh\left(\frac{\boldsymbol{X}_{p}^{T}\boldsymbol{W}_{jp}\boldsymbol{J}}{\sqrt{d}}\right).
\end{equation}

\noindent
where $\boldsymbol{W}_{jp} \in \mathbb{R}^{L \times L}$ represents the learnable weight matrix across P and joint P-V features. Similarly, the joint correlation matrix of the viseme features is given by Equation \eqref{eq:10}.
\begin{equation}\label{eq:10}
 \boldsymbol{M}_{v}= \tanh\left(\frac{\boldsymbol{X}_{v}^{T}\boldsymbol{W}_{jv}\boldsymbol{J}}{\sqrt{d}}\right).
\end{equation}

After calculating the joint correlation matrix, the attention weights of P and V modalities are estimated. Since the joint correlation matrix ($\mathbb{R}^{dp \times d}$) differs from the feature dimension of the corresponding modalities ($\mathbb{R}^{L \times d_{p}}$), we calculate the attention weights using different learnable weight matrices corresponding to the features of the individual modalities. 
For the P modality, the joint correlation matrix $\boldsymbol{W}_{p}$ and the corresponding $\boldsymbol{X}_{p}$ are combined using the learnable weight matrices $\boldsymbol{W}_{mp}$ and $\boldsymbol{W}_{p}$, respectively.
\begin{equation}\label{eq:11}
 \boldsymbol{H}_{p} = \operatorname{ReLU}(\boldsymbol{W}_{p}\boldsymbol{X}_{p} + \boldsymbol{W}_{mp} \boldsymbol{M}_{p}^{T}).
\end{equation}

\noindent
where $\boldsymbol{W}_{mp} \in \mathbb{R}^{k \times d}$, $\boldsymbol{W}_{p} \in \mathbb{R}^{k \times L}$, and $\boldsymbol{H}_{p}$ denotes the attention maps of modality P. Similarly, the attention map $\boldsymbol{H}_{v}$  of modality V is obtained as Equation \eqref{eq:12}:
\begin{equation}\label{eq:12}
 \boldsymbol{H}_{v} = \operatorname{ReLU}(\boldsymbol{W}_{v}\boldsymbol{X}_{v} + \boldsymbol{W}_{mv} \boldsymbol{M}_{v}^{T}).
\end{equation}

\noindent
where $\boldsymbol{W}_{mv}  \in \mathbb{R}^{k \times d}$, $\boldsymbol{W}_{v}  \in \mathbb{R}^{k \times L}$.

Finally, the attention features of P and V modalities are obtained as follows:
\begin{equation}\label{eq:13}
  \boldsymbol{Att}_{p}  = \boldsymbol{W}_{hp}\boldsymbol{H}_{p} + \boldsymbol{X}_{p},
\end{equation}
\begin{equation}\label{eq:14}
  \boldsymbol{Att}_{v}  = \boldsymbol{W}_{hv}\boldsymbol{H}_{v} + \boldsymbol{X}_{v}.
\end{equation}
where $\boldsymbol{W}_{hp} \in \mathbb{R}^{k \times L}$ and $\boldsymbol{W}_{hv} \in \mathbb{R}^{k \times L}$ denote the learnable weight matrices, respectively. 

%In this way, we can allow the features of one modality to calibrate the features of the other modality while maintaining the correlation between the different modalities. 
Connecting $\boldsymbol{X}_{att,p}$ and $\boldsymbol{X}_{att,v}$ further yields the P-V feature representation, which is given by Equation \eqref{eq:15}:
\begin{equation}\label{eq:15}
  \boldsymbol{X}_{Att}
= \left\lbrack \boldsymbol{Att}_{p}; \boldsymbol{Att}_{v} \right\rbrack.
\end{equation}

\noindent
\textbf{Co-correlation Guided Representation Alignment Module.} After multimodal feature fusion, we present a Co-correlation Guided Representation Alignment Module (CGRA) to align heterogeneous audio-visual multimodal data. CGRA aims to strengthen the correlation between two modalities as much as possible by making the correlation matrix between the network outputs close to the identity matrix as Equation \eqref{eq:16}. Representation alignment can reduce the modality gap better to explore the intrinsic complementarity of the two modalities.
\begin{equation}\label{eq:16}
  {Loss}_{cor} \triangleq \sum_{i}(1-\boldsymbol{C}_{ii})^{2} + \lambda\sum_{i} \sum_{j \neq i} \boldsymbol{C}_{ij}^{2}.
\end{equation}

\noindent
where $\lambda$ is a positive constant weighing the importance of the first and second terms of the loss.%where C is the order of the cross-correlation matrix obtained between two identical networks outputs along the batch size.
\begin{equation}\label{eq:17}
\boldsymbol{C}_{ij} \triangleq \frac{\sum_{b}\boldsymbol{Att}_{p_{i}}^{b}\boldsymbol{Att}_{v_{j}}^{b}}
{\sqrt{\sum_{b}\boldsymbol{{Att}_{p_{i}}^{b}}^{2}}\sqrt{\sum_{b}\boldsymbol{{Att}_{v_{j}}^{b}}^{2}}}.
\end{equation}

\noindent
where $b$ is the batch samples and $\boldsymbol{C}_{ij}$ is a square matrix of the same size as the output of the network with values between -1 (perfectly anti-correlated) and 1 (perfectly correlated).

\begin{table*}[t]
	\centering
	\caption{Fine-tuning the model for videos with different compression rates on the FF++ dataset and testing the performance at that compression rate. The best results are shown in bold.}
	\label{tablefine-tun}
	\tabcolsep=1.5pt
	\renewcommand\arraystretch{0.9}
	\begin{tabular*}{\linewidth}{@{\hspace{\tabcolsep}}@{\extracolsep{\fill}}ccccccccc@{\hspace{\tabcolsep}}}
		\toprule
		\multirow{3}{*}{Method} & \multirow{3}{*}{Modality} &\multirow{3}{*}{Pretrain} &  \multicolumn{3}{c}{\scriptsize Video-level ACC(\%)} & \multicolumn{3}{c}{\scriptsize Video-level AUC(\%)} \\
		\cmidrule{4-9}
		& & & Raw & HQ & LQ & Raw & HQ & LQ \\
		\midrule
		Patch-based\cite{chai2020makes} & visual & - & 98.9 & 92.6 & 79.1 & 99.8 & 97.1 & 78.3 \\
		CNN-aug\cite{wang2020cnn} & visual & - & 98.7 & 96.9 & 81.6 & 99.8 & 99.1 & 86.9 \\
		Xception\cite{rossler2019faceforensics++} & visual & - & 98.9 & 97.0 & 89.1 & 99.8 & 99.3 & 91.4 \\
		Two-branch\cite{masi2020two}& visual & - & - & - & - & - & 99.1 & 91.0 \\
		Face X-ray\cite{li2020face}& visual & Sup-BI & 99.1 & 78.4 & 34.2 & 99.8 & 97.6 & 77.3 \\
		LipForensics\cite{haliassos2021lips} & visual & Sup-LRW & 98.9 & 98.0 & 94.2 & 99.4 & \textbf{99.7} & 96.1 \\
		Self-LipForensics\cite{haliassos2021lips} & visual & Sup-LRW & 98.6 & 96.5 & 88.4 & 99.8 & 99.3 & 94.8 \\
		RECCE\cite{cao2022end} & visual & - & 99.1 & 97.1 & 91.0 & 99.8 & 99.3 & 95.0 \\
		\midrule
		%Emotions Don't Lie\cite{mittal2020emotions}& audio-visual & -&-&-&-&-&-&-\\
		Joint A-V Detection\cite{zhou2021joint} & audio-visual &-& 98.6 &98.0&\textbf{95.8}&99.3&99.0&91.4\\
		AVoiD-DF\cite{yang2023avoid}& audio-visual & -& 99.0 &98.2&93.9&\textbf{99.9}&99.2&93.5\\
		SST\cite{zhao2022self} & audio-visual & Self-Sup & \textbf{99.2} & \textbf{98.5} & 93.5 & 99.7 & 99.6 & 95.7 \\
		
		VFD\cite{cheng2022voice} & audio-visual& Self-Sup & 98.3 & 94.2 & 90.1 & 99.4 & 96.5 & 89.6 \\

		\midrule
		NPVForensics (Ours) & audio-visual & Self-Sup & \textbf{99.2} & \textbf{98.5} & 95.6 & \textbf{99.9} & \textbf{99.7} & \textbf{96.4} \\
		\bottomrule
	\end{tabular*}
\vspace{-0.3cm}
\end{table*}

\subsection{Pre-training and Fine-tuning}
\label{Pre-train}

\noindent
\textbf{Pre-training protocol}. We utilize three loss functions to update the parameters of the pre-training model. 

1) Based on non-critical phoneme-viseme evolutionary consistency, we design the evolutionary consistency loss $L_{EC}$.

2) Based on audio-visual homogeneity \cite{cheng2022voice}, we use unsupervised contrast loss InfoNCE between audio and face features.

3) Alignment loss $L_{cor}$ for the CGRA module. 

Thus, the total loss in the pre-training stage is calculated as Equation \eqref{eq:18}:
\begin{equation}\label{eq:18}
  L_{pre}=L_{EC}+L_{Info}+L_{cor}.
\end{equation}

\noindent
\textbf{Fine-tuning on Deepfake Dataset}. %Benefiting from the pre-training of relatively large scale real videos, the model has good learning capability.
We perform a downstream specific Deepfake detection task using the pre-trained model.
In the fine-tuning and testing phase, a concat operation is added to the pixel-level global representation and the fused phoneme-viseme representation to obtain the final discriminative feature. Fine-tuning is performed using Deepfake datasets with a binary classifier. The classification uses cross-entropy loss. The cross-entropy loss is defined as Equation \eqref{eq:19}:
\begin{equation}\label{eq:19}
  L_{ce} = - \frac{1}{N}\sum_{i = 1}
  \resizebox{0.9\width}{!}{$\left\lbrack y_{i} \cdot \log(P_{i}) + (1 - y_{i}) \cdot \log(1 - P_{i}) \right\rbrack$}.
\end{equation}
where $y_{i}$ is the one-hot vector of the ground-truth of video, $P_{i}$ is the predicted probability vector. Finally, the total loss function of the overall framework in the fine-tuning stage is given by the following Equation \eqref{eq:20}:
\begin{equation}\label{eq:20}
  L =L_{pre}+wL_{ce}.
\end{equation}
where $w$ is the scale factor, which we set to 1.

%%%%%%%%%%%%%
%Experiment
%%%%%%%%%%%%%

\section{Experiments}

\noindent
\textbf{Auxiliary Datasets.} We utilize the $VoxCeleb2$ dataset \cite{chung2018voxceleb2} and part of the $AVSpeech$ dataset \cite{ephrat2018looking} for pre-training, with a total of 2500,000 speech video clips. More than 100 consecutive frames with a trackable face are included in each video clip. We divide them in an 8:1:1 ratio into training, validation, and test sets.

\noindent
\textbf{Deepfake Datasets.} We utilize Deepfake datasets as follows. 1) \textbf{FaceForensics++(FF++)} \cite{rossler2019faceforensics++} consists of 1000 real videos and 4000 fake videos, generated using two face-swapping methods, Deepfakes and FaceSwap, and two face reenactment methods, Face2Face and Neural Textures. We utilize the mildly compressed dataset version (c23) unless stated otherwise. Real audio extracts the Mel-spectrum directly from the audio in the existing dataset containing the sound. Fake audio converts the text to audio by the $ text\ to\  speech $ (TTS) algorithm\cite{jia2018transfer} and then extracts the converted audio from the Mel-spectrum. 2) \textbf{FaceShifter (FSh)} \cite{li2019faceshifter} and 3) \textbf{Celeb-DF-v2} \cite{li2020celeb} with 518 test videos. 4) \textbf{DeeperForensics (DFo)} \cite{jiang2020deeperforensics} are state-of-the-art face-swapping methods that applied to the real videos of FF++; we use the test videos according to the FF++ split. 5) \textbf{Deepfake Detection Challenge Dataset (DFDC)}\cite{dolhansky2020deepfake} contains 23,654 real videos and 104,500 fake videos from 960 identities. Since some videos in that dataset are mixed with the camera holder's voice, we altered 6000 real videos and 32,200 fake videos as the fine-tuned set, while 1700 real videos and the corresponding 5810 fake videos were used as the test set. 6) For \textbf{FakeAVCeleb} \cite{khalid2021fakeavceleb}, 500 real videos and 19,500 fake videos were included, where we used 391 real videos and the corresponding 16,869 fake videos as the fine-tuning set and the rest as the test set. 7) We analyze lip-sync Deepfakes created through the use of three synthesis techniques: \textbf{Audio-to-Video (A2V)}\cite{suwajanakorn2017synthesizing} and \textbf{Text-to-Video (T2V)} \cite{fried2019text} in which only short utterances are manipulated (T2V-S), and Text-to-Video in which longer utterances are manipulated (T2V-L). We selected over 450 real and 1940 fake videos as the fine-tuning set, while 100 real videos and the corresponding 428 fake videos were used as the test set. 

\noindent
\textbf{Evaluation metrics.} Following the evaluation metrics from previous work on Deepfake detection, we evaluate the detection performance using video-level ROC-AUC scores (AUC) and binary classification accuracy (ACC).

\noindent
\textbf{Implementation Details.} We implemented our model using the Pytorch toolkit. The basic learning rate of the finetune model is $1\times{10}^{-3}$ (for the finetune exclusive module) and $5\times{10}^{-6}$ (for the pretrain and finetune shared module) using the AdamW optimizer. Parameters for pre-training and fine-tuning are updated, and the respective learning rates decay gradually with the training process, with weights decaying by $1\times{10}^{-2}$. The model is trained on four NVIDIA RTX 3090 Ti with a mini-batch size of 64. We used 12 Transformer blocks with 8 head self-attention in the identity feature extractor. The input face images were tuned to $224 \times 224$, and the voice clips were represented as $80,000 \times 32$ sound spectrograms (5 seconds of video) as the voice clips were divided into 16,000 small windows per second.

\subsection{Evaluation on FaceForensics++}

The FF++ dataset is currently the most widely used dataset for Deepfake detection. We use four forgery methods to fine-tune our model. In addition, the FF++ videos have three compression rates, uncompressed (Raw), slightly compressed (HQ), and heavily compressed (LQ). As the compression level increases, some texture features are lost and the detection difficulty increases.

Table~\ref{tablefine-tun} shows the video-level AUC and accuracy at various compression rates compared to other techniques. %The Xception\cite{rossler2019faceforensics++}, CNN-aug\cite{wang2020cnn}, Patch-based\cite{chai2020makes} and Two-branch\cite{masi2020two} methods are not pretrained. 
Face X-ray \cite{li2020face} generates training data by blending real faces and being supervised with blending boundary regression. LipForensics \cite{haliassos2021lips} is pretrained in a supervised manner on the Lipreading in the Wild (LRW) dataset \cite{2016Lip}. To parallel compare our self-supervised method to the supervised one, we also utilize the same network structure and pretrain it on LRW with the same training strategy of LipForensics in a supervised manner, the model is denoted as LipForensics*. Joint A-V Detection\cite{zhou2021joint} propose a novel visual$/$auditory Deepfake joint detection task to exploit the intrinsic synchronization between the visual and auditory modalities. SST \cite{zhao2022self} is a self-supervised Transformer based on audio-visual contrast learning, which also uses a large amount of real data for pre-training. VFD \cite{cheng2022voice} utilizes audio-visual homogeneity for self-supervised contrast learning. AVoiD-DF\cite{yang2023avoid} uses a dual-stream temporal-spatial encoder and a multimodal joint decoder to learn intrinsic relations between audio and visual.%RealForensics \cite{haliassos2022leveraging} is a two-stage detection approach using a pre-training strategy that successively performs representation learning and multi-task forgery detection. 
We notice that NPVForensics performs better than the supervised model. Furthermore, it performs well on original videos and high compression rate videos.

\subsection{Cross-manipulation Generalization}

The Deepfake detector can detect unseen forgeries in the training set, i.e., it has a strong generalization capability, which is vital in real-world applications. In this experiment, we evaluate the generalization ability of the cross-manipulation method on the FF++ HQ dataset. We test the results on one manipulation method and train on the other three data types. The results are shown in Table~\ref{tableff++}.
\begin{table}[t]
	\centering
	\caption{Cross-manipulation evaluation. Video-level AUC (\%) when testing on each forgery type of FF++ HQ after training on the remaining three. The best results are shown in bold.}
	\label{tableff++}
	\small
	\renewcommand\arraystretch{0.95}
	\begin{tabular}{cccccc}
		\toprule
		\multirow{2.5}{*}{Method} & \multicolumn{4}{c}{Train on remaining three} & \multirow{2.5}{*}{\textbf{Avg}} \\
		\cmidrule{2-5}
		& DF & FS & F2F & NT &  \\
		\midrule
		Patch-based\cite{chai2020makes} & 94.0 & 60.5 & 87.3 & 84.8 & 81.7 \\
		CNN-aug\cite{wang2020cnn} & 87.5 & 56.3 & 80.1 & 67.8 & 72.9 \\
		Xception\cite{rossler2019faceforensics++} & 93.9 & 51.2 & 86.8 & 79.7 & 77.9 \\
		Face X-ray\cite{li2020face} & 99.7 & 90.1 & 99.2 & 98.1  & 96.8 \\
		Self-LipForensics\cite{haliassos2021lips} & 97.8 & 90.5 & 98.0 & 96.9 & 95.8 \\
			RECCE\cite{cao2022end} & 98.2 & 88.9 & 92.1 & 88.5 & 91.9 \\
		\midrule
		Emotions Don't Lie\cite{mittal2020emotions} &94.5 &89.3 & 86.9 & 93.7 & 91.1\\
		Joint A-V Detection\cite{zhou2021joint} &97.7 &90.5 &\textbf{99.7} &97.3&96.3\\
		SST\cite{zhao2022self} & 94.5 & 91.9 & 98.3 & 96.4 & 95.3 \\
	
		VFD\cite{cheng2022voice} & 96.2 & 86.3 & 89.6 & 94.2 & 91.6 \\
		AVoiD-DF\cite{yang2023avoid}&97.3&94.7&94.5&95.1&95.4\\
		\midrule
		\textbf{NPVForensics (Ours)} & \textbf{99.8} & \textbf{96.2} & 99.4 & \textbf{98.6} & \textbf{98.5} \\
		\bottomrule
	\end{tabular}
\vspace{-0.3cm}
\end{table}
Our method outperforms methods such as Xception and RECCE that use only visual modality, demonstrating the effectiveness of introducing audio modality to assist visual modality for Deepfake detection. NPVForensics achieves better results compared to methods using audio-visual multimodality. It suggests that NPVForensics can effectively capture fine-grained audio-visual inconsistencies and explore correlations in non-critical phoneme-viseme regions with better generalization.

\subsection{Generalization Across Datasets}

In realistic scenarios, the forgeries distribution is diverse, and different Deepfake datasets have large domain gaps. Therefore, domain generalization ability is an essential criterion for Deepfake detection models. In this section, we fine-tune our method on the FF++ dataset and report the AUC scores for tests performed on Celeb-DF-v2 (CDF), DFDC, FaceShifter HQ (FSh), and DeeperForensics (DFo), respectively.

In Table~\ref{tablecross-dataset}, our approach achieves advanced performance in generalizing to CDF, FSh and DFo datasets. The performance of FSh and DF is undoubtedly superior to CDF and DFDC, with AUCs of 98.2\% and 98.6\%, respectively. It is likely because both datasets and FF++ use the same original video.
NPVForensics achieves an average AUC of 92.5\% on the four datasets, outperforming the Face X-ray\cite{li2020face} and Lipforensics\cite{haliassos2021lips} by 11.3\% and 5.4\%, respectively. Multimodal Deepfake detection methods generally obtained better results than other unimodal methods. NPVForensics outperforms AVoiD-DF\cite{yang2023avoid} in the cross-dataset evaluation by 2.3\%. There are two possible reasons to explain. One is that the self-supervised learning-based approach learns adequate audio-visual representations in a large number of real videos, which can significantly improve the generalization of the model; the second is that our approach focuses more on the regions that the forger overlooks. Compared with the AVoiD-DF\cite{yang2023avoid}, our method is more likely to tap into subtle inconsistencies and intrinsic complementarity features and thus achieves promising results on unseen Deepfakes.

\begin{table}[t]

\centering
\caption{Generalizability across datasets. Video-level AUC (\%) tests on CDF, DFDC, FaceShifter HQ (FSh), and DFo. Here, The best results are shown in bold (Train on FF++).}
\label{tablecross-dataset}
\small
\renewcommand\arraystretch{0.95}
\begin{tabular}{cccccc}
\toprule
Method & CDF & DFDC & FSh & DFo & \textbf{Avg} \\
\midrule
Xception\cite{rossler2019faceforensics++} & 73.7 & 70.9 & 72.0 & 84.5 & 75.3 \\
Face X-ray\cite{li2020face} & 79.5 & 65.5 & 92.8 & 86.8 & 81.2 \\
CNN-GRU\cite{sabir2019recurrent} & 69.8 & 68.9 & 80.8 & 74.1 & 73.4 \\
LipForensics\cite{haliassos2021lips} & 82.4 & 73.5 & 95.9 & 96.6 & 87.1 \\
Self-LipForensics\cite{haliassos2021lips} & 83.6 & 73.6 & 96.1 & 97.0 & 87.6 \\
RECCE\cite{cao2022end} & 78.5 & 69.1 & 84.9 & 88.7 & 80.4 \\
\midrule
Emotions Don't Lie\cite{mittal2020emotions}&82.1&84.4&96.0&96.3&89.7\\
Joint A-V Detection\cite{zhou2021joint} & 83.9 &76.5&96.9 &95.8&88.3\\
SST\cite{zhao2022self} & 84.2 & 74.5 & 97.8 & 97.3 & 88.4 \\
VFD\cite{cheng2022voice} & 80.7 & \textbf{85.1} & 85.9 & 84.3 & 84.0 \\
AVoiD-DF\cite{yang2023avoid}&86.2 &82.3 &95.7&96.5&90.2\\

\midrule
\textbf{NPVForensics (Ours)} & \textbf{88.4} & 84.9 & \textbf{98.2} & \textbf{98.6} & \textbf{92.5} \\
\bottomrule
\end{tabular}
\vspace{-0.3cm}
\end{table}

\begin{table}[ht]
\vspace{-0.1cm}
	\centering
	\caption{Generalizability across datasets. Video-level AUC (\%) tests on A2V, T2V-L and T2V-S. The best results are shown in bold (Train on FF++).}
	\label{tablecross-dataset-A2V}
	\small
	\renewcommand\arraystretch{0.95}
	\begin{tabular}{ccccc}
		\toprule
		Method & A2V & T2V-L & T2V-S & \textbf{Avg} \\
		\midrule
		Xception\cite{rossler2019faceforensics++} & 81.6 & 63.9 & 75.3 & 73.6  \\
		Face X-ray\cite{li2020face} & 77.2 & 60.3 & 81.4 & 73.0  \\
		CNN-GRU\cite{sabir2019recurrent} & 80.1 & 66.3 & 82.2 & 76.2  \\
		LipForensics\cite{haliassos2021lips} & 84.4 & 74.2 & 86.6 & 81.7 \\
		Self-LipForensics\cite{haliassos2021lips} & 84.6 & 74.3 & 87.1 & 82.0  \\
		RECCE\cite{cao2022end} & 69.1 & 66.3 & 78.2 & 71.2  \\
		\midrule
		Joint A-V Detection\cite{zhou2021joint} &80.5 & 69.8 & 80.2 &76.8\\
		
		SST\cite{zhao2022self} & 89.6 & 77.4 & 86.7 & 84.6  \\
		
		VFD\cite{cheng2022voice} & 67.8 & 60.9 & 65.4 & 64.7  \\
		PVM\cite{agarwal2020detecting} & 94.6 & 79.7 & 74.1 & 82.8  \\
		AVoiD-DF\cite{yang2023avoid}& 85.1 &78.3 &84.4 &82.6\\
		\midrule
		\textbf{NPVForensics (Ours)} & \textbf{97.1} & \textbf{81.3} & \textbf{91.9} & \textbf{90.1}  \\
		\bottomrule
	\end{tabular}
\vspace{-0.4cm}
\end{table}

\begin{table*}[h]
	\centering
	\caption{Robustness to common degradations. The models are fine-tuning with FakeAVCeleb datasets. AUC(\%) scores at five intensity levels are averaged for each degradation type. The best results are shown in bold.}
	\label{tablerobustness}
	\small
	\renewcommand\arraystretch{0.9}
	\begin{tabular*}{\linewidth}{@{\hspace{\tabcolsep}}@{\extracolsep{\fill}}ccccccccc@{\hspace{\tabcolsep}}}
		\toprule
		Method & Saturation & Contrast & Block & Noise & Blur & Pixel & Compress & Avg \\
		\midrule
		Patch-based\cite{chai2020makes} & 84.3 & 74.2 & 98.8 & 50.0 & 54.4 & 56.7 & 53.4 & 67.5 \\
		CNN-aug\cite{wang2020cnn} & 99.3 & 99.1 & 95.2 & 54.7 & 76.5 & 91.2 & 72.5 & 84.1 \\
		Xception\cite{rossler2019faceforensics++} & 99.3 & 98.6 & 98.7 & 53.8 & 60.2 & 74.2 & 62.1 & 77.3 \\
		Face X-ray\cite{li2020face} & 97.6 & 88.5 & \textbf{99.1} & 49.8 & 63.8 & 88.6 & 55.2 & 77.5 \\
		LipForensics\cite{haliassos2021lips} & \textbf{99.9} & \textbf{99.6} & 87.4 & 73.8 & 93.1 & 95.6 & 95.6 & 92.5 \\
		Self-LipForensics\cite{haliassos2021lips} & 99.8 & 99.0 & 87.1 & 73.7 & 94.2 & 93.7 & 93.2 & 91.6 \\
		SST\cite{zhao2022self} & 98.3 & 98.2 & 89.3 & 64.8 & 94.0 & 89.2 & 86.7 & 88.6 \\
		RECCE\cite{cao2022end} & 91.7 & 91.2 & 88.4 & 65.3 & 87.3 & 83.9 & 89.7 & 85.4 \\
	
		\midrule
		\textbf{NPVForensics (Ours)} & 99.7 & 99.2 & 97.9 & \textbf{79.7} & \textbf{94.3} & \textbf{96.2} & \textbf{97.6} & \textbf{94.9} \\
		\bottomrule
	\end{tabular*}
	\vspace{-0.1cm}
\end{table*}

\begin{figure*}[ht]
	\centering
	\includegraphics[width=0.85\linewidth]{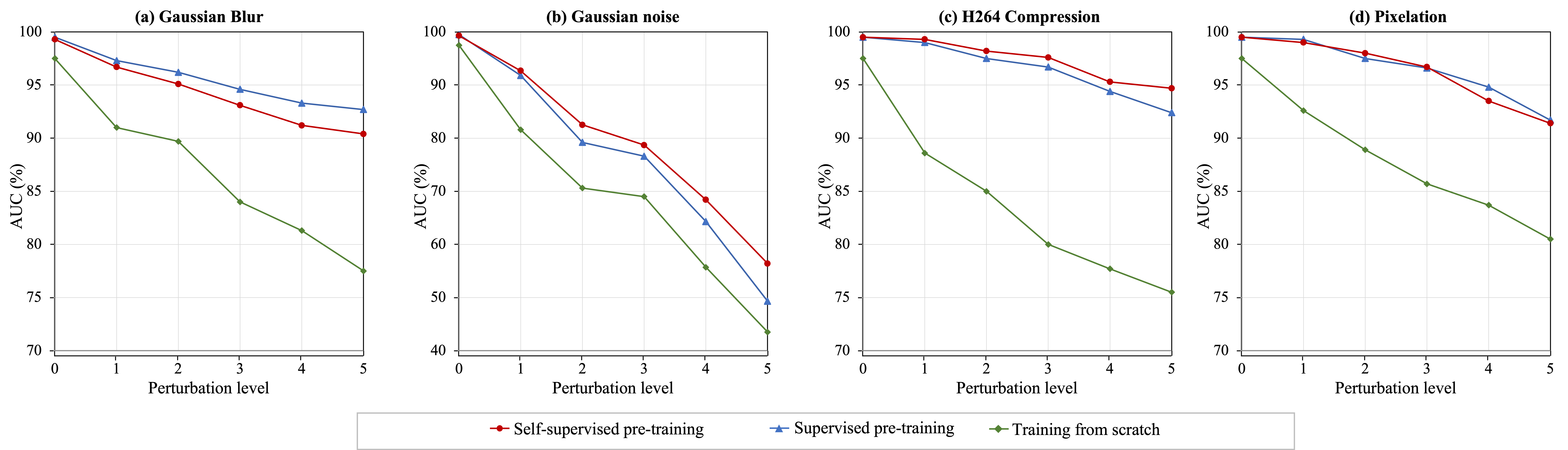}
	\caption{The vertical axis represents the video-level AUC score, and the horizontal axis represents the severity level of perturbations. The four perturbations are Gaussian blur, block-level distortion, H264 compression, and pixelation decomposition. We compare the robustness of the model with different pre-training strategies.}
	\label{fig:6}
	\vspace{-0.5cm}
\end{figure*}

Table~\ref{tablecross-dataset-A2V} shows the generalization performance on the three lip-sync Deepfake datasets: A2V, T2V-L and T2V-S. These datasets have a more realistic tampering effect by considering the critical phoneme-viseme coordination during the generation process and thus convey a challenge for the existing Deepfake detection techniques.

As shown in Table~\ref{tablecross-dataset-A2V}, our method has good detection performance for Deepfakes generated by such synthetic techniques, with an average AUC exceeding SST\cite{zhao2022self} by 5.5\% and higher than PVM\cite{agarwal2020detecting} by 7.3\%. The reason is that those novel forgeries are calibrated at the critical phoneme-viseme region, so when PVM\cite{agarwal2020detecting} and SST\cite{agarwal2020detecting} extracts salient features for Deepfake detection, it extracts the calibrated critical phoneme-viseme features, leading to a certain degree of misclassification. We note that PVM\cite{agarwal2020detecting} outperforms T2V-L over T2V-S due to the higher ratio of calibrated critical phonemes and visemes in T2V-S, and thus T2V-S is more challenging to detect for PVM\cite{agarwal2020detecting}.
NPVForensics average AUC higher than AVoiD-DF\cite{yang2023avoid} by 7.5\%, demonstrating that NPVForensics extraction of non-critical phoneme-viseme has significant advantages for detecting lip-sync-driven Deepfakes. Our method is an effective detection tool to tackle the more realistic Deepfake videos in the future.

\subsection{Robustness Evaluation}

In addition to possessing the capability of good generalization, the detector should be robust to common video degradation. We follow LipForensics\cite{haliassos2021lips} to evaluate robustness to unseen perturbations. As in \cite{haliassos2021lips}, we finetune on FakeAVCeleb using grayscale clips without any augmentation other than horizontal flipping and random cropping to avoid any intersection between training and testing temporal perturbations. The perturbations proposed in our paper are variations of saturation, contrast, block masking, Gaussian noise, blurring, pixelation, and video compression. Each perturbation type is applied to five different intensity levels. Table~\ref{tablerobustness} shows the average AUC for all intensity levels for each degradation type. When comparing the results with state-of-the-art methods, NPVForensics outperforms LipForensics and texture-based techniques. It is due to the fact that we pretrained the feature extractor LFA-ST on a significant amount of real videos to mine the essential distinctions between real videos and Deepfake videos. The evolutionary consistency and intrinsic complementarity of non-critical phoneme-viseme can provide sufficient discrimination basis for detection. Additionally, it shows that high-level semantic features are more robust to perturbations.

We focus on the robustness of the model under different pretraining strategies. The three pretraining strategies are Train from Scratch, Self-supervised pretrain and Supervised pretrain. Degradations of the definition include adding block-wise distortions, Gaussian blurring, JPEG compression, and H264 video compression. We can draw conclusion from Fig.~\ref{fig:6} that our self-supervised pretraining strategy has better robustness than supervised pretraining.

\subsection{Visualization}
\begin{figure*}[ht]
	\centering
	\includegraphics[width=0.6\linewidth]{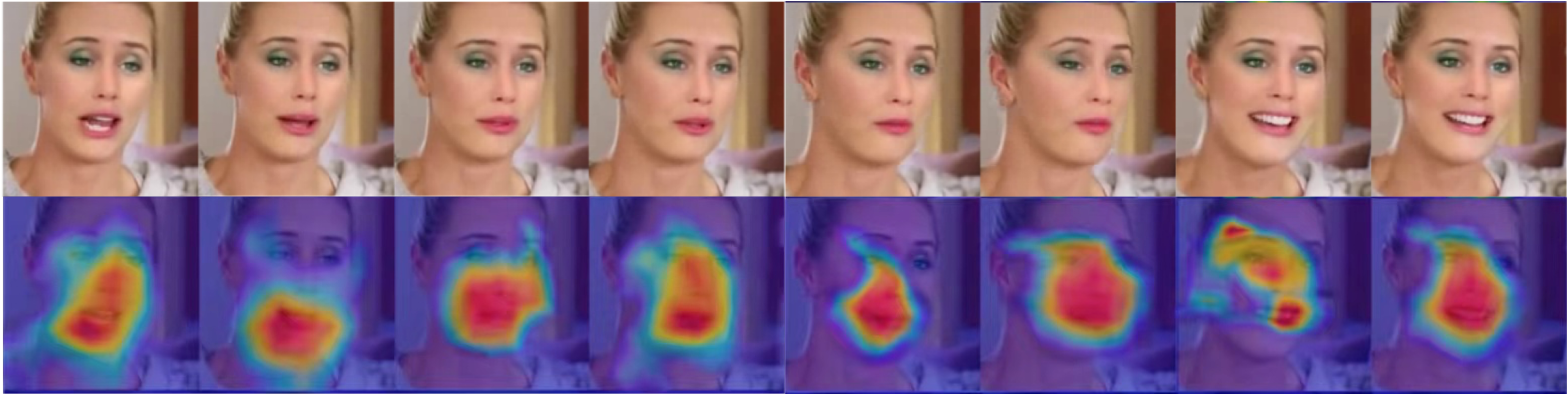}
	\caption{Visualization of the manipulated regions of sequential video frames from FakeAVCeleb. The manipulated regions are concentrated in the lips area, where warmer color indicates a higher probability of forgery.}
	\label{fig:frad-cam}
	\vspace{-0.3cm}
\end{figure*}

%The following example is a Deepfake video from FakeAVCeleb, as shown in Fig.~\ref{fig:frad-cam}. Its facial identity information has not been tampered with, and only the visemes have been manipulated to match the fake audio. This instance is a Deepfake video generated by the Lip-sycn technique, which is misclassified as an actual video in methods Self-LipForensics, RECCE, and VFD. To explore which region our method focuses on when encountering unseen forgeries, we use Grad-CAM to generate heat maps. Our method effectively focuses on irrelevant audio-visual areas and correctly detects the authenticity of the video.

The following example is a Deepfake video from FakeAVCeleb, generated by Lip-sync technology. To explore which region our method focuses on when dealing with unseen forgeries, we employ Grad-CAM to generate a heat map, as shown in Fig.~\ref{fig:frad-cam}. Its identity information is not tampered with; only the visual information is tampered with to match the fake audio. It is misclassified as real video in methods such as RECCE\cite{cao2022end}, Self-LipForensics\cite{haliassos2021lips}, VFD\cite{cheng2022voice}, SST\cite{zhao2022self} and AVoiD-DF\cite{yang2023avoid}. The reasons are that the faked video has few visual artifacts, the identity information is not changed, and the critical phoneme-viseme is calibrated. Our method accurately finds the forgery region and effectively exploits the correlation of non-critical phoneme-viseme regions for forgery detection. As shown in Fig.~\ref{fig:frad-cam}, our method focuses on the local region (lip movement) manipulated by the forger.

\subsection{Ablation Experiments}
\begin{table}[b]
\vspace{-0.4 cm}
\centering
\caption{Framework ablation. Accuracy scores (\%) of FaceShifter HQ (FSh), DeeperForensics (DFo) and FakeAVCeleb after training on the FF++ dataset. The best results are shown in bold.}
\label{tableframe-ablation}
\small
\renewcommand\arraystretch{1.0}
\tabcolsep=3pt
\begin{tabular}{ccccc}
\toprule
\multirow{2.5}{*}{Method} & \multicolumn{3}{c}{FakeAVCeleb (audio-visual)}&\textbf{Avg}  \\
\cmidrule{2-4}
 & Real-Fake & Fake-Fake & Fake-Real \\
\midrule
w/o visual stream & - & 87.7 & 87.9 & - \\
w/o facial stream & 93.5 & 95.7 & 95.9 & 95.0 \\
w/o phoneme stream & 89.9 & 93.3 & 93.6 & 92.3\\
w/o LFA, only ST & 94.8 & 95.3 & 93.9 & 94.7\\
ViT & 95.0 & 94.9 & 93.9 & 94.6 \\
w/o CAFM & 91.7 & 92.8 & 94.3 & 92.9 \\
w/o CGRA & 92.5 & 93.3 & 94.1 & 93.3\\
\midrule
\textbf{NPVForensics (Ours)} & \textbf{96.7} & \textbf{97.8} & \textbf{96.3} & \textbf{96.9}\\
\bottomrule
\vspace{-0.6 cm}
\end{tabular}
\end{table}

\noindent
\textbf{Framework Ablation.} In this section, we introduce ablation to understand the factors contributing to the performance of NPVForensics. In Table~\ref{tableframe-ablation}, we ablate different components and check its generalization performance on FSh, DFo and FakeAVCeleb after training on FF++. We make the following observations. The performance decreases slightly when there is no facial stream. It demonstrates that pixel-level features also play a positive role in Deepfake detection. Then, eliminating the audio stream, the performance on FakeAVCeleb decreased by 4.6\%. The reason is that part of the data in this dataset is only manipulated with the lips region. Almost no visual artifacts result in the performance degradation of only visual features. That demonstrates that audio-visual inconsistency is a reasonable basis for Deepfake detection. Without the CAFM module, the AUC decreases by about 4.0\%. Without CGRA to perform feature alignment, the AUC decreases by about 3.6\%. It suggests that the cross attention fusion module with the co-covariate guided representation alignment module is meaningful. Notably, the performance comparison of ViT, ST, and LFA-ST highlights the effectiveness of our proposed LFA module. The addition of the LFA module aggregates local features, combined with the global features extracted from the Swin Transformer, helps to mine the weak correlation of non-critical phonemes and visemes (relative to critical phoneme-viseme) and facilitates the detection of fine-grained forged talking faces.

%Additionally, we compare the number of parameters of LFA-ST and ST. Parameters of Swin Transformer and LFA-ST are 34.2M and 34.4M, respectively. The number of parameters is the same. After adding LFA module, the performance will be improved. In addition, we also use CPU and GPU for image prediction of the FF++ dataset. The number of prediction times and parameters are shown in Table~\ref{tablepram}. The results show that LFA not only ensures the accuracy, but also improves the computational efficiency with only a small increase in the number of parameters.
%
%
%
%
%\begin{table}[ht]
%\centering
%\caption{Parameters and inference of the two methods.}
%\label{tablepram}
%\small
%\renewcommand\arraystretch{1.2}
%\begin{tabular}{cccc}
%\toprule
%
%\multirow{2.5}{*}{Method} & \multirow{2.5}{*}{Parameters} & \multicolumn{2}{c}{Inference} \\
%\cmidrule{3-4}
% &  & CPU & GPU \\
%\midrule
%ST\cite{liu2022video} & 34.2M & 49.3 ms & 5.8 ms \\
%LFA-ST & 34.4M & \textbf{45.9 ms} & 6.0 ms \\
%\bottomrule
%\end{tabular}
%\end{table}

\noindent
\textbf{Effect of Pretraining Dataset Scale.} 
\begin{figure}[t]
	\centering
	\includegraphics[width=0.8\linewidth]{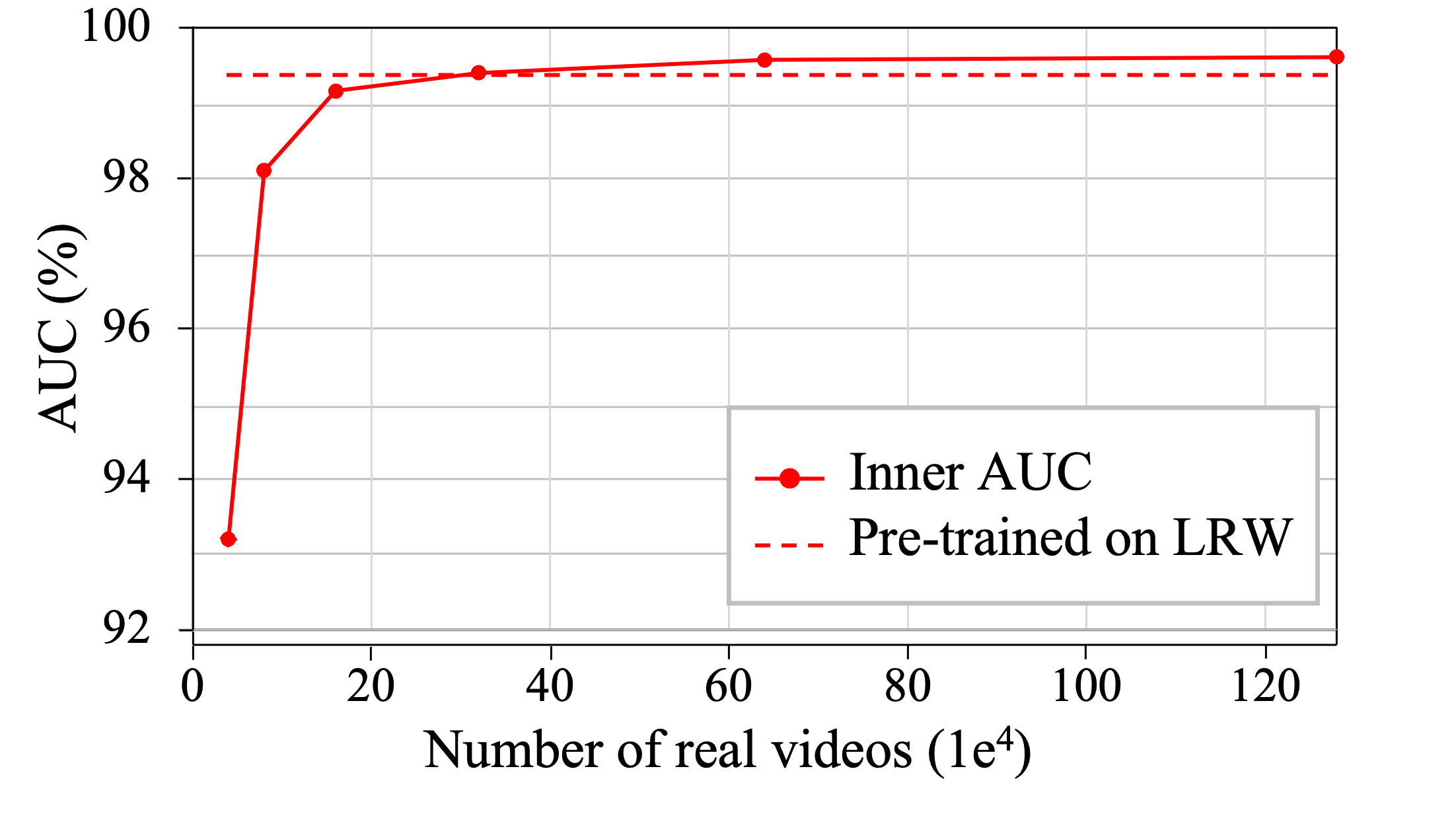}
	\includegraphics[width=0.8\linewidth]{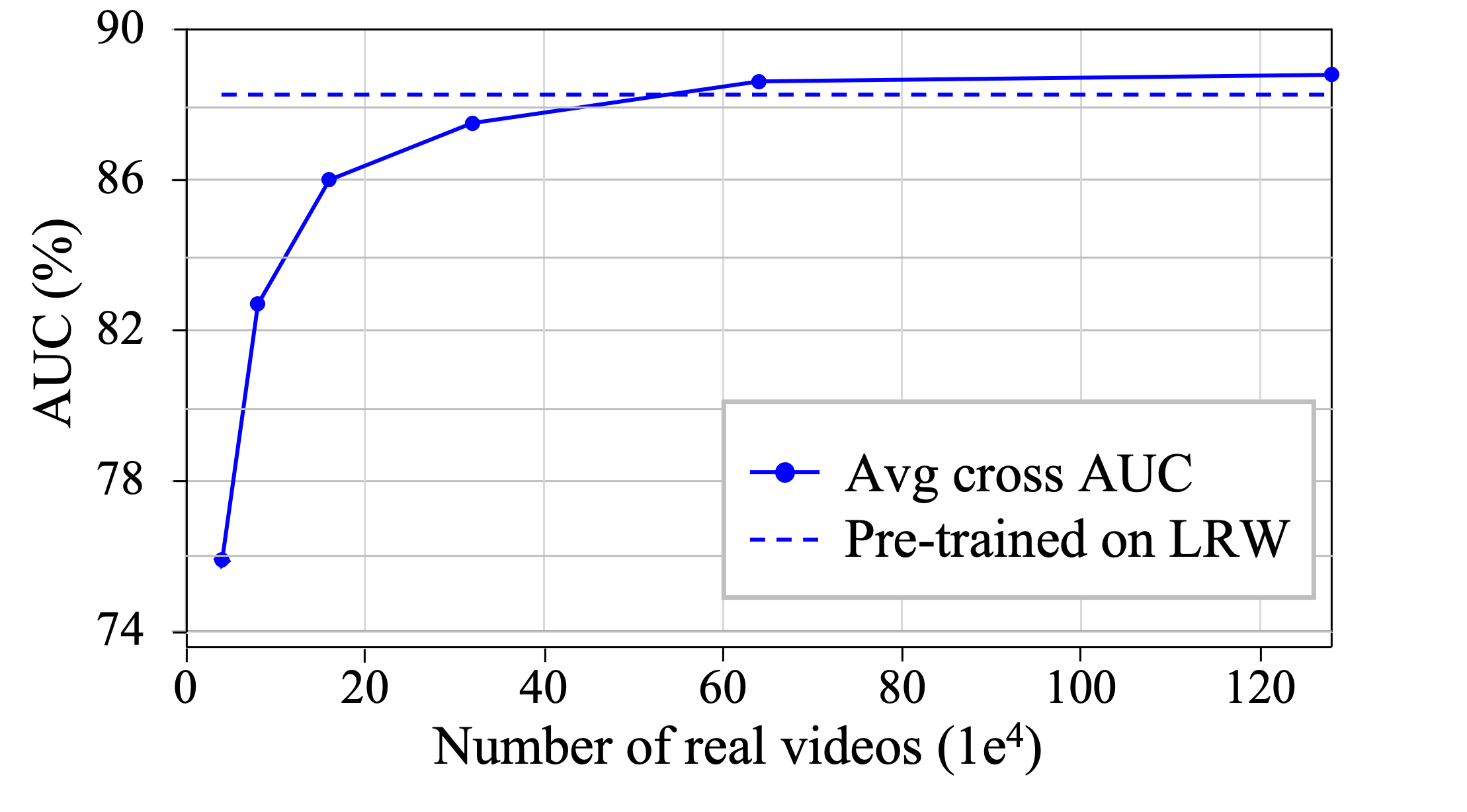}
	\caption{The relation between pretrained data scale and AUC score. The above chart shows the relation between the pre-training data scale and the test AUC on the FF++ HQ dataset. The below chart shows the relation between the pretraining data scale and the average cross-dataset AUC fine-tuning on the FF++ dataset.}
	\label{fig:7}
	\vspace{-0.cm}
\end{figure}
One of the advantages of self-supervised learning is that it utilizes a large number of easy-to-collect real face videos for pre-training without the necessity of expensive manual annotations. Comparatively speaking, the labelled dataset LRW for the lip reading task contains about 500,000 video clips of 500 words with precise boundary annotations. It spends a lot of effort on annotation. Therefore, Deepfake detection using self-supervised learning would be an attractive direction. In this section, we investigate how the scale of the pretrained data affects the model's performance. The relationship curves are shown in Fig.~\ref{fig:7}. The horizontal dotted line represents the AUC score of the pretrained model on the LRW dataset of 500,000 video clips. The solid line presents the evolution of the AUC score as the scale of the self-supervised pretrained data increases. The solid red line indicates the AUC scores for testing on the FF++HQ dataset. The solid blue line indicates the average cross-dataset AUC scores fine-tuned on the FF++ dataset.

The curves in Fig.~\ref{fig:7} validate that models pre-trained with more real videos can improve accuracy within and across datasets. When pretrained with 320,000 video clips, the performance outperforms the same network pre-trained with 500,000 labelled video clips from the LRW dataset. In conclusion, we demonstrate that larger-scale pre-training data helps improve the performance of our model. However, more extensive training data inevitably leads to higher computational resource requirements, which may be a limitation.

\section{Conclusion}

This paper proposes a novel cue for effective Deepfake detection that concentrates on mining phoneme-viseme correlations by focusing on non-critical phoneme-viseme regions that forgers cannot perfectly calibrate. To achieve this goal, we propose a new framework termed NPVForensics. Considering that the features of non-critical phoneme regions are relatively subtle, we designed LFA-ST as the backbone for extracting features so as to maximize the retention of global and local information. We utilize the phoneme and viseme features for phoneme-viseme evolutionary consistency analysis. In addition, to effectively exploit the intrinsic complementarity between audio and visual modalities, we design a phoneme-viseme awareness module to fuse audio-visual features and introduce a co-correlation guided representation alignment module to reduce the modalities gap. We employ a self-supervised strategy to pretrain on large-scale real videos and fine-tune the framework on Deepfake datasets. In practical applications, the accuracy and generalization of the model can be improved by scaling up the model and data scale. We have demonstrated that our method achieves strong cross-manipulation generalization and robustness to common corruptions.

%%%%%%%%% REFERENCES

{\small
\bibliographystyle{IEEEtran}}
\bibliography{refs}

\end{document}